\documentclass{article}

\usepackage{PRIMEarxiv}

\usepackage[utf8]{inputenc} 
\usepackage[T1]{fontenc}    
\usepackage{hyperref}       
\usepackage{amsfonts} 
\usepackage{amssymb}
\usepackage{url}            
\usepackage{booktabs}       
\usepackage{amsfonts}       
\usepackage{nicefrac}       
\usepackage{microtype}      
\usepackage{lipsum}
\usepackage{fancyhdr}       
\usepackage{graphicx}       
\graphicspath{{media/}}     

\title{Machine learning approaches for localized lockdown during COVID-19: a case study analysis
}

\author{
  Sara Malvar$^*$ \\
  Microsoft \\
  \texttt{saramalvar@microsoft.com} \\
   \And
  Julio Romano Meneghini$^*$\\
  $^*$Department of Mechanical Engineering, Escola Politécnica \\
  University of S\~{a}o Paulo \\
}

\begin{document}
\maketitle

\begin{abstract}
At the end of 2019, the latest novel coronavirus Sars-CoV-2 emerged as a significant acute respiratory disease that has become a global pandemic. Countries like Brazil have had difficulty in dealing with the virus due to the high socioeconomic difference of states and municipalities. Therefore, this study presents a new approach using different machine learning  and deep learning algorithms applied to Brazilian COVID-19 data. First, a clustering algorithm is used to identify counties with similar sociodemographic behavior, while Benford's law is used to check for data manipulation. Based on these results we are able to correctly model SARIMA models based on the clusters to predict new daily cases. The unsupervised machine learning techniques optimized the process of defining the parameters of the SARIMA model. This framework can also be useful to propose confinement scenarios during the so-called second wave. We have used the 645 counties from São Paulo state, the most populous state in Brazil. However, this methodology can be used in other states or countries. This paper demonstrates how different techniques of machine learning, deep learning, data mining and statistics can be used together to produce important results when dealing with pandemic data. Although the findings cannot be used exclusively to assess and influence policy decisions, they offer an alternative to the ineffective measures that have been used.  
\end{abstract}


\section{Introduction}
In a few weeks, the pandemic outbreak of the COVID-19 virus changed the lives of people around the world. In many countries, the pace of infection, the lack of natural immunological defences in humans and the high mortality rates due to the disease have challenged the capacity of the health system and forced almost all daily activities in urban centers to be temporarily shut down in order to ensure the required social distancing. In the last six months, a considerable number of works was published considering studies related to the COVID-19 evolution through the world \cite{Gorbalenya, Anderson, Biswas, Zhou, Zhao}. The case of Brazil has been considered in some of these papers \cite{Crokidakis}, but none of them report several machine learning applications using the same data. Our interest in this work is to study the evolution of COVID-19 in the most populous Brazilian state, namely São Paulo and its 645 counties, as Brazil has become the global epicenter of the outbreak in the last few months. 
    
For countries like Brazil, which present great social and economic inequality, the construction of public policies to face the world pandemic of COVID-19 brings significant challenges \cite{MARMOT}. In these vulnerable countries, a number of problems are encountered when trying to create quarantine policies. In most extreme cases, lockdown practices have been applied. These lead to stagnation in several sectors of the economy, as a result of the abrupt drop in consumption and production. It may increase unemployment and a significant reduction in the population's income, further aggravating the living condition of people in situations of social vulnerability. On the other hand, the lack of restrictions on people's mobility can lead to a significant increase in the demand for equipment and medical staff, sometimes incompatible with the capacity of the national system. Despite high transmissibility, the disease spectrum is diverse, ranging from asymptomatic cases to extremely severe conditions \cite{CHENG}. The outcome of the disease is directly related to the socio-economic conditions of the population \cite{CAMARA}.

The lockdown policies seem to attract growing popularity across a vast number of governments. In some countries such as Italy and France, drastic lockdown measure imposed a national-level of mobility restriction. On other countries, such as Brazil, a more moderate restriction applies regional levels. Many countries are usually reluctant to directly impose a lockdown because of the economic consequences \cite{RAHMAN}. 

Currently, ongoing efforts have been made to develop novel diagnostic approaches and predict new cases numbers using machine learning algorithms \cite{ALIMADADI}. Neural network classifiers, for example, were developed for large-scale screening of COVID-19 patient based on their distinct respiratory patterns \cite{WANG}. Similarly, for automatic detection and monitoring of COVID-19 patients over time,
a deep learning-based analysis system of thoracic CT images was developed for \cite{GOZES}. Machine learning has also been used to predict epidemic trends in COVID-19 \cite{ZHU, BOCCALETTI}.

At this time, most of the models are not deployed enough to show their real-world operation and have localized specificities, which makes their generalization difficult. However, they are still up to the mark to tackle the SARS-CoV-2 epidemic. 

This work aims to analyze the spread of the pandemic in the state of São Paulo in three stages. In the first stage, a thoughtful data mining process is used to obtain 85 sociodemographic variables from 645 counties.The variables are used to cluster the counties considering their similarities. This segmentation may allow the government to take more effective measures concerning social distancing and quarantine, based on the cities characteristics. The second analysis aims to identify the possibility of underreporting considering Benford's Law. This analysis aims not only to identify possible manipulations of data but also a possible flattening of the contagion curve. Finally, a seasonal ARIMA models based on the segments obtained in the first stage is proposed.

\section{Methods}
\label{sec:headings}
\subsection{Data acquisition}

This study mainly uses two datasets: one time series that contains information about the COVID-19 pandemic and one related to the sociodemographic situation of all 645 counties of São Paulo state. The last is a single, unified dataset with 85 columns, containing several counties profile characteristics. Socioeconomic data were obtained from several sources. The raw data and dictionary are available on supplementary material as table 1 and also on the authors Github \cite{GITHUB}.

The daily data related to the number of new cases and new deaths by COVID-19 for counties were downloaded from the Fundação Sistema Estadual de Análise de Dados (SEADE) \cite{SEADE}, translated as State System of Data Analysis Foundation. We processed the original data in two significant ways. First, we normalized counties names to use them as merging key since there were minor differences in the name of the countries among the datasets. Second, we used the most up-to-date variable whenever the different datasets provided the same variable for different times. 

\subsection{Clustering}

Clustering algorithms are generally used in an unsupervised fashion.  They are presented with a set of data instances that must be grouped according to some notion of similarity. In our case, we used the socioeconomic data to cluster the counties. This segmentation can make it easier for the government to make localized quarantine decisions, which have been used in the UK \cite{LI}.

We have applied K-means clustering  \cite{MACQUEEN}, a method commonly used to partition a dataset into $k$ groups automatically. It proceeds by selecting $k$ initial cluster centers and then iteratively refining them as follows:

\begin{itemize}
    \item Each instance $d_i$ is assigned to its closest cluster center.
    \item Each cluster center $C_j$ is updated to be the mean of its constituent instances.
\end{itemize}

The algorithm converges when there is no further change in the assignment of instances to clusters. The number of variables used was reduced after calculating the correlation between them to avoid multicollinearity. Variables that showed a correlation greater than $0.9$ had their pairs removed. Data was also scaled between 0 and 1 to prevent the variance between the order of magnitude of the variables from affecting the result. 

To determine the ideal amount of clusters, we have used the elbow rule. The basic idea of the elbow rule is to use a square of the distance between the sample points in each cluster and the centroid of the cluster to give a series of K values. The sum of squared errors (SSE) is used as a performance indicator and smaller values indicate that each cluster is convergent. When the number of clusters is set to approach the number of real clusters, SSE shows a rapid decline. When the number of clusters exceeds the number of real clusters, SSE will continue to decline, but it will quickly become slower \cite{YUAN}. Using this technique, we defined $9$ clusters, which was also confirmed by the silhouette score \cite{Ogbuabor}.

\subsection{Benford Distribution}

To identify possible manipulations in the data, we have used the Benford distribution.  It is shown that if numbers are taken from an
exponential distribution, they immediately obey Benford's Law (BL) \cite{NEWCOMB}. As a result, we could assume that the number of infections and deaths of COVID-19 follow BL when the disease growth curve approaches exponential distribution. This was an empirically  discovered pattern for the frequency distribution of first digits in many real-life datasets \cite{BOYAU}. Previous studies have shown that Benford's Law is applicable to self-reported toxic emissions data \cite{MARCHI}, tax audit \cite{NIGRINI}, election data \cite{GAMERMANN}, stock markets \cite{AUSLOOS} and religion \cite{MIR}. 

Aiming to understand the validity of SARS-CoV-2 data on São Paulo's counties, the second part of this study compare the cases official data and Benford's Law \cite{NEWCOMB, PACINI}. Analyzing data from countries affected by the pandemic, it is possible to observe patterns that are common to all. Despite the existence of several peculiarities such as territorial extension, population density, temperature, season, degree of underreporting and social discipline to comply with isolation measures, the virus has not suffered radical mutations since its appearance in China and the general parameters of infectivity and lethality are similar between countries.

On the other hand, data from São Paulo's counties are a notable exception, showing behavior that differs between themselves. Benford's Law defines that in many collections of numbers from real-life data or mathematical tables, the significant digits are not uniformly distributed; they are heavily skewed toward the smaller digits. More precisely, the significant digits in many datasets obey a very particular logarithmic distribution:

\begin{equation}
    P = \log_{10} \left( 1 + \frac{1}{D} \right),
\end{equation}

\noindent where $P$ is the probability of first (non‐zero) digit $D$ occurring ($D = 1, ..., 9)$. For example, the real numbers $123.0$ and $0.016$ both have $D = 1$, and the digit law suggests that numbers beginning with a $1$ will occur about $30\%$ of the time in nature, while those with the first digit of $2$ will occur about $17\%$ of the time, and so on down to first digits of $9$ occurring about $4\%$ of the time. From a statistical standpoint, a Borel probability measure $P$ on $\mathbb{N}$ is Benford if:

\begin{equation}
P({x \in \mathbb{N}: S(x) \leqslant u}) = \log u
\end{equation}

\noindent for all $u \rightarrow [1,10)$, where $S$ is the significant of a real number is its coefficient when it is expressed as a floating-point \cite{LEE}.

The ramifications of the digit rule are important, as not only is the distribution not universal, meaning that the digit frequencies are not unique, but, to be valid, they must also be retained irrespective of the data units and their source. As a consequence, the fundamental property of real-world estimation is inferred. In order to determine the degree of variation between the observed and predicted distribution of the first digit from BL, we used the chi-square test, which can be calculated as: 

\begin{equation}
    \chi^2 = \sum_{i=0}^9 \frac{(n_i - p_i)^2}{p_i},
\end{equation}

\noindent where $n_i$ is the observed frequency in each bin in the observed data and $p_i$ is the expected frequency based on Benford’s distribution.

\subsection{Seasonal ARIMA Model}

Seasonal autoregressive integrated moving average models (SARIMA) provide another approach to time series forecasting. It aims to describe the autocorrelations in the data. However, defining the parameters of ARIMA is not always easy. In addition, when dealing with seasonal data, which is the case, we must include additional seasonal terms. We will present a brief presentation of Seasonal ARIMA model below, but the reader is referred to a comprehensive time series analysis text, such as Brockwell and Davis \cite{DAVIS} or Fuller \cite{FULLER} for further details.

Let $Y' = (Y_1, Y_2, ... ,Y_n)$ be a time series of data. The Seasonal ARIMA model with $S$ observations per period, detoned by \textit{SARIMA(p,q,d)(P,Q,D)S}, is given by:

\begin{equation}
    \Phi(L^S)\Phi(L)(1-L)^d (1-L^S)^D Y_t = \theta(L^S) \theta (L) E_t,
\end{equation}

\noindent where $L$ is the lag operator given by $L^k = Y_{t-k} / Y_y$, $\Phi(L) = 1 - \Phi_1 L^1 - \Phi_2 L^2 - ... \Phi_p L^p$ is an autoregressive (AR) polynomial function of order $p$ with vectors of coefficients $\Phi' = (\Phi_1, \Phi_2, ..., \Phi_p)$, $\theta(L) = 1 + \theta_1L^1 + \theta_2 L ^2 + ... + \theta_q L^q$ is a moving average (MA) polynomial of order $q$ with vectors of coefficient $\theta' = (\theta_1, \theta_2, ..., \theta_q)$ and $\Phi(L^S) = 1 - \Phi_{s,1}L^S - \Phi_{S,2}^{2S} - ... - \Phi_{S,P}L^{PS} $ and $\theta(L^S) = 1 - \theta_{s,1}L^S - \theta_{S,2}^{2S} - ... - \theta_{S,Q}L^{QS} $ are seasonal polynomial functions of order $P$ and $Q$, respectively, that satisfy the stationarity and invertibility conditions. The order of differencing needed to stationarize the series is $d$, $D$ is the number of seasonal differences and \textit{E}$_t$ are error terms assumed to be independent identically distributed random variables sampled from a distribution with zero mean. The model was applied using different combination of parameters for the cases of each city and the model with the lowest Akaike information criterion (AIC) value for each dataset was computed. 

To define ARIMA parameters, we must understand if the variance grows with time and if a variance-stabilizing transformation or differencing is needed. Then, we might use autocorrelation function (ACF) to measure the amount of linear dependence between observations in a time series that are separated by a lag \textit{p}, and the partial autocorrelation function (PACF) to determine how many autoregressive terms \textit{q} are necessary and inverse autocorrelation function (IACF) for detecting over differencing, we can identify the preliminary values of autoregressive order \textit{p}, the order of differencing \textit{d}, the moving average order \textit{q} and their corresponding seasonal parameters \textit{P}, \textit{D} and \textit{Q}.

\section{Results}

\subsection{Counties Clustering}

After clustering, the mean sociodemographic characteristics of each cluster of counties can be found in Table 2 of the supplementary material. 

Neves Paulista and Itaporanga are represented by cluster 0. It represents cities with a low rate of urbanization and birth rates, in addition to a high rate of inhabitants per car. There are few health professionals, and the population density is low. In general, they are small cities, with the majority of jobs between livestock, agriculture and local commerce.

Cluster 1 includes ABC Paulista (São Bernardo do Campo, Diadema and São Caetano), Mogi Guaçu, Sorocaba and Rio Claro. Most formal jobs are in the industry, and the human development index of the municipalities is high. Per capita income is the highest of all clusters, but the population density is also high. As they are cities with a lot of industrial activities and many factories, the management of employees must be done with great care, because in many of these industries the processes cannot stop.

Cluster 2 presents the characteristics of the major counties. It has the capital: the city of São Paulo. In conjunction, several other municipalities such as Santos, Bertioga, Guarujá, Presidente Prudente and São José do Rio Preto. These cities have a high degree of urbanization, high coefficient per thousand inhabitants of nurses, nursing assistants and other health professionals. On the other hand, the population density is extremely high, and the number of inhabitants per vehicle is over 3. This probably means that people usually live in houses with more than one individual and use public transport, which are risk factors for COVID-19 spreading. Also, the average level of sanitary sewage service is low.

On the other hand, cluster 3, where we may find Bofete and Cesário Lange, for example, there is a high percentage of jobs associated with agriculture and livestock. The level of urbanization is slightly lower, and there is a much lower number of doctors per 1000 inhabitants. These are cities with a high index of human development and with good quality of life, in addition to a reasonably low population density, which avoids agglomeration. The mortality rate for the population aged 15 to 34 is also minimal.

This is a different scenario from cluster 4 that contains Cunha, Joanópolis and Nazaré Paulista. In this case, the percentage of jobs related to agriculture is quite relevant, and the per capita income and population density are also lower than cluster 0. However, as they are cities with tourist potential, there is a reasonable number of health professionals. They are probably cities with insufficient evidence of agglomeration but prepared to treat those affected.

Cluster 5, which includes cities like Bauru, Guarulhos and Taboão da Serra, has an extremely high population density. GDP per capita is relatively high, and most jobs are in the wholesale, retail trade and industry. 

Cities like Marinópolis and Bento de Abreu, Holambra and Ibiúna, with a high level of jobs in livestock and agriculture are in cluster 6 and 7. However, the low population density is also represented in the low number of health professionals and per capita income. Both clusters have a high degree of basic sanitation. The big difference between the two is that, in addition to jobs in livestock, fishing and agriculture, cluster 6 has a reasonable percentage of jobs in the industry, while cluster 7 has a reasonable percentage of jobs in the trade.

Vinhedo, Arujá and Cajamar, represented by cluster 8, are near the capital and present high employment rate in the industry and population density. Despite having a reasonable number of nurses and nursing assistants, it does not have a reasonable number of doctors, which can be problematic.

\begin{figure}[h!]
    \centering
      \includegraphics[scale=1.7]{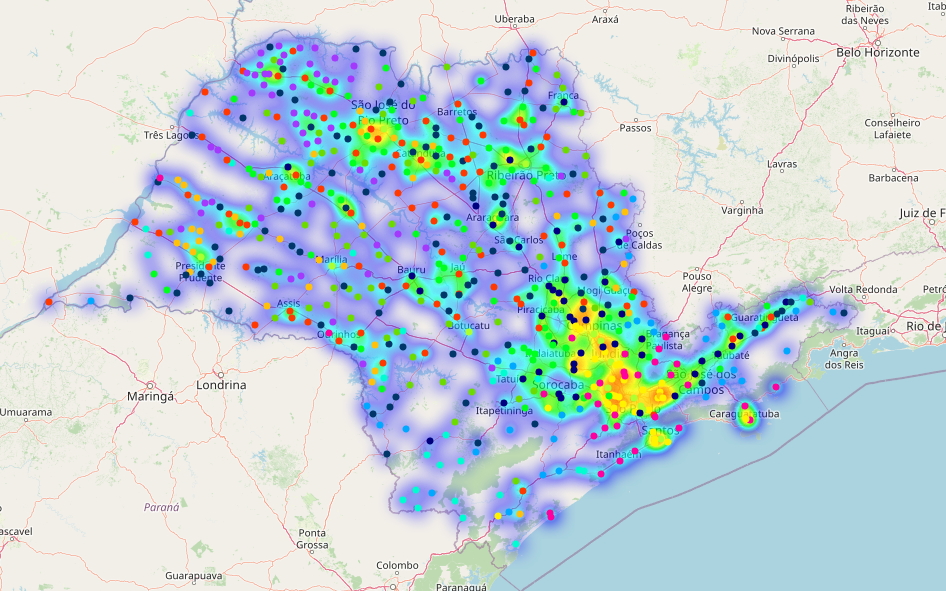}
      \label{fig:mean and std of net14-wavelet}
    \caption
    {Map of the state of São Paulo. Each color represents a cluster. There is also a heat map behind that shows the focus of COVID-19 cases until December 13 of 2020. The intensity goes from blue to green and then yellow and red.} 
    \label{map}
\end{figure}
  
Figure \ref{map} shows the map of São Paulo state and the counties with their respective cluster. This method demonstrates that there are clear sociodemographic clusters of counties throughout São Paulo state. Additionally, because the clustering in this paper was based on sociodemographic indicators that are possibly associated with the outcomes of interest, one might hypothesize that members of a given cluster would benefit from similar public health interventions, making these clusters excellent groups for learning exchange. 
 
\begin{table}[h!]
\centering
\begin{tabular}{|c|c|c|}
\hline
\textbf{Cluster} & \textbf{Cases} & \textbf{Fatalities} \\ \hline
0 & 118.28 & 2.54 \\ \hline
1 & 409.26 & 8.39 \\ \hline
2 & 742.44 & 17.38 \\ \hline
3 & 5654.86 & 189.84 \\ \hline
4 & 4681.26 & 137.54 \\ \hline
5 & 261.44 & 5.56 \\ \hline
6 & 15021.26 & 565.97 \\ \hline
7 & 904.53 & 23.17 \\ \hline
8 & 433.6 & 11.07 \\ \hline
\end{tabular}
\caption{Table showing the average number of cases and deaths for each cluster.}
\label{casos}
\end{table}

We see from table \ref{casos} that the clusters with the highest cases and fatalities are 6, 3, 4 and 7. These are the places with the highest population density. The major problem is that counties of clusters 0, 1, 2, 3, 5, 8 account for less than 10\% of the total state GDP. That means that potential restrictions could be imposed on areas that now have high COVID-19, but economic aid plans must be implemented.
  
\subsection{Benford Law and Data Consistency}

Some uncertainty arises from the fact that being a new disease for which biochemical tests are not widely available, the cases surveillance is essential. Decisions based on evidence need useful quality data. Idrovo \cite{IDROVO, GOMES} used Newcomb-Benford's law to assess the performance of surveillance systems during influenza A (H1N1) pandemic. This method has been used on other occasions, including Zika \cite{MANRIQUE}, and dengue epidemic in American countries. This law states that for a determined set of numbers, those whose first digit is 1 will appear more frequently (30.103\%) than those beginning with other digits, following in order from 2 to 9 (17.609\%, 12.494\%, 9.691\%, 7.918\%, 6.695\%, 5.799\%, 5.115\%, and 4.576\%, respectively) \cite{HILL}. 

In this section, we investigate the use of the first digit distribution to detect potential anomalies in reported Covid-19 data of cities from São Paulo state. All data are recorded daily from February 25 of 2020. Depending on the municipal public policies, the number of reported cases and deaths are not standardized, which has given rise to massive debates. However, it is not the intention of this paper to present or discuss these policies but to identify possible statistical discrepancies that may even make it harder to forecast new cases.

Considering the existence of 645 municipalities in the state of São Paulo and the possibility that many had few cases of COVID-19, we limited the analysis to municipalities with more than 5000 cases, culminating in 25 samples. We perform  $\chi^2$ goodness-of-fit test, Kolmogorov-Smirnov test \cite{MATHER} and Mean Absolute Deviation (MAD) test to judge the adequacy of the Newcomb-Benford distribution. In addition, the Z-scores were computed, considering a significance of 5\% (p-value = 0.05). The results are shown in table \ref{tabela1}.

\begin{table}[]
\begin{center}
\begin{tabular}{cccccc}
\hline
\textbf{Municipality}          &$\ \mathbf{chi^2}$  & \textbf{Test} & \textbf{KS}       & \textbf{Test} & \textbf{MAD}      \\ \hline
Campinas              & 37.7765 & Fail            & 0.1779 & Fail      & 0.015000 \\ \hline
Osasco                & 28.4385    & Fail            & 0.1490   & Fail      & 0.037803 \\ \hline
Sorocaba              & 47.9916 & Fail            & 0.2068 & Fail      & 0.050021 \\ \hline
Ribeirão Preto        & 12.9550 & Pass            & 0.0948 & Pass      & 0.024839 \\ \hline
Jundiaí               & 16.6665 & Fail            & 0.0720 & Pass      & 0.029562\\ \hline
São José do Rio Preto & 22.3186 & Fail            & 0.1040 & Fail      & 0.033295 \\ \hline
São José dos Campos   & 27.5192  & Fail           & 0.1440 & Fail     & 0.037979 \\ \hline
Praia Grande          & 69.3898 & Fail            & 0.1330 & Fail      & 0.045155 \\ \hline
Guarujá               & 71.0776 & Fail            & 0.2204 & Fail      & 0.057966 \\ \hline
Bauru                 & 5.2733 & Pass            & 0.0646 & Pass      & 0.017032 \\ \hline
Mogi das Cruzes       & 32.2669  & Fail            & 0.1131 & Fail      & 0.032081 \\ \hline
Taboão da Serra       & 62.3477 & Fail            & 0.1093 & Fail     & 0.043406 \\ \hline
Araçatuba             & 53.4126  &Fail            & 0.1422 & Fail      & 0.041060 \\ \hline
Barueri               & 53.4126 & Fail            & 0.1422 & Fail      & 0.041060 \\ \hline
Hortolândia           & 31.8301 & Fail           & 0.0718 & Pass      & 0.030611 \\ \hline
São Paulo             & 69.9758 & Fail            & 0.2121 & Fail      & 0.047126 \\ \hline
Santos                & 33.1445 & Fail            & 0.1825 & Fail      & 0.040545 \\ \hline
São Bernardo do Campo & 40.7767 & Fail            & 0.1912 & Fail      & 0.042488 \\ \hline
Guarulhos             & 56.0600 & Fail            & 0.2363 & Fail      & 0.053395 \\ \hline
Santo André           & 44.2709 & Fail            & 0.1940 & Fail     & 0.047464 \\ \hline
Piracicaba            & 26.1314 & Fail          & 0.1139 & Fail      & 0.036287 \\ \hline
Diadema               & 74.7318 & Fail            & 0.2126 & Fail      & 0.053901 \\ \hline
São Vicente           & 104.4507 & Fail            & 0.1777 & Fail      & 0.065583 \\ \hline
Carapicuíba           & 44.3929 & Fail            & 0.1000 & Fail      & 0.042429 \\ \hline
Cubatão               & 76.7173 & Fail            & 0.1767 & Fail      & 0.057376 \\
\hline
\end{tabular}
\caption{Results from $\chi^2$, Kolmogorov-Smirnov test and Mean Absolute Deviation (MAD) test to judge the  adequacy of the Newcomb-Benford distribution from 25 counties.}
\label{tabela1}
\end{center}
\end{table}

The interpretations of our results are based on the assumption that variations from Newton-Benford law are indicative of data manipulation. 
Indeed, many studies in macroeconomic, accounting, finance and forensic analysis show that human intervention and data manipulation creates datasets that violate Newcomb-Benford law. 

\begin{figure}[h!]
\begin{center}
 \includegraphics[width=\textwidth]{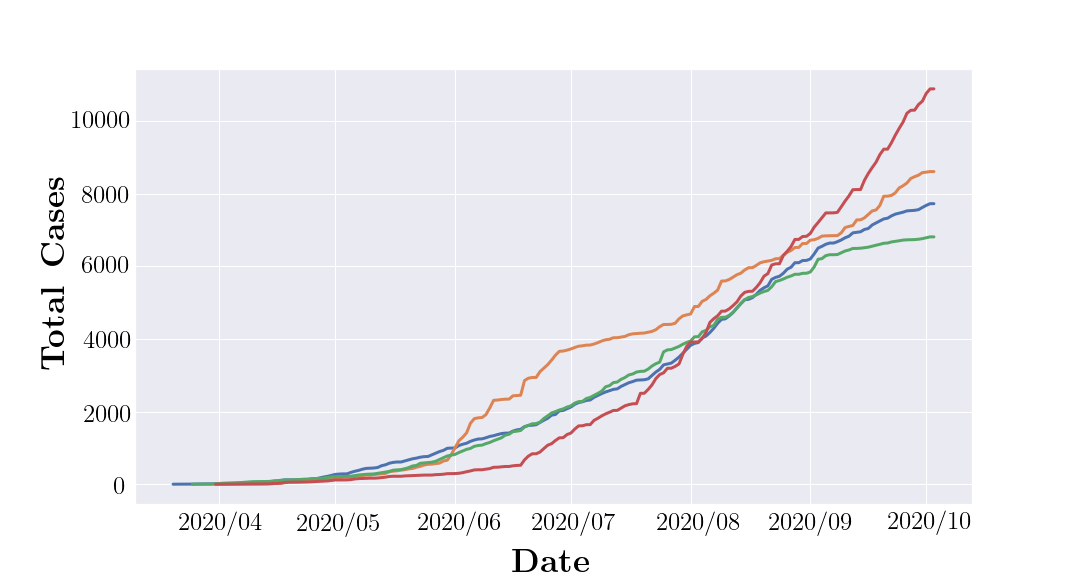}
 \caption{Total cases of Guarujá (orange line), Mogi das Cruzes (blue line), Taboão da Serra (green line) and Bauru (red line).}
   \label{casos50}
\end{center}
\end{figure}

To facilitate the analysis, a comparison between four specific cases is made: Guarujá, Mogi das Cruzes, Taboão da Serra and Bauru, which passes both Kolmogorov and $\chi^2$ tests. Thus, figure \ref{casos50} shows the total number of cases as a function of the time of these four counties.

On the other hand, according to Lee et al. \cite{LEE} if the current control interventions are successful and we flatten the curve (i.e., we slow the rate below an exponential growth rate), then the number of infections or deaths will not obey Benford's law. 

\begin{figure}
    \centering
      \includegraphics[scale=1.3]{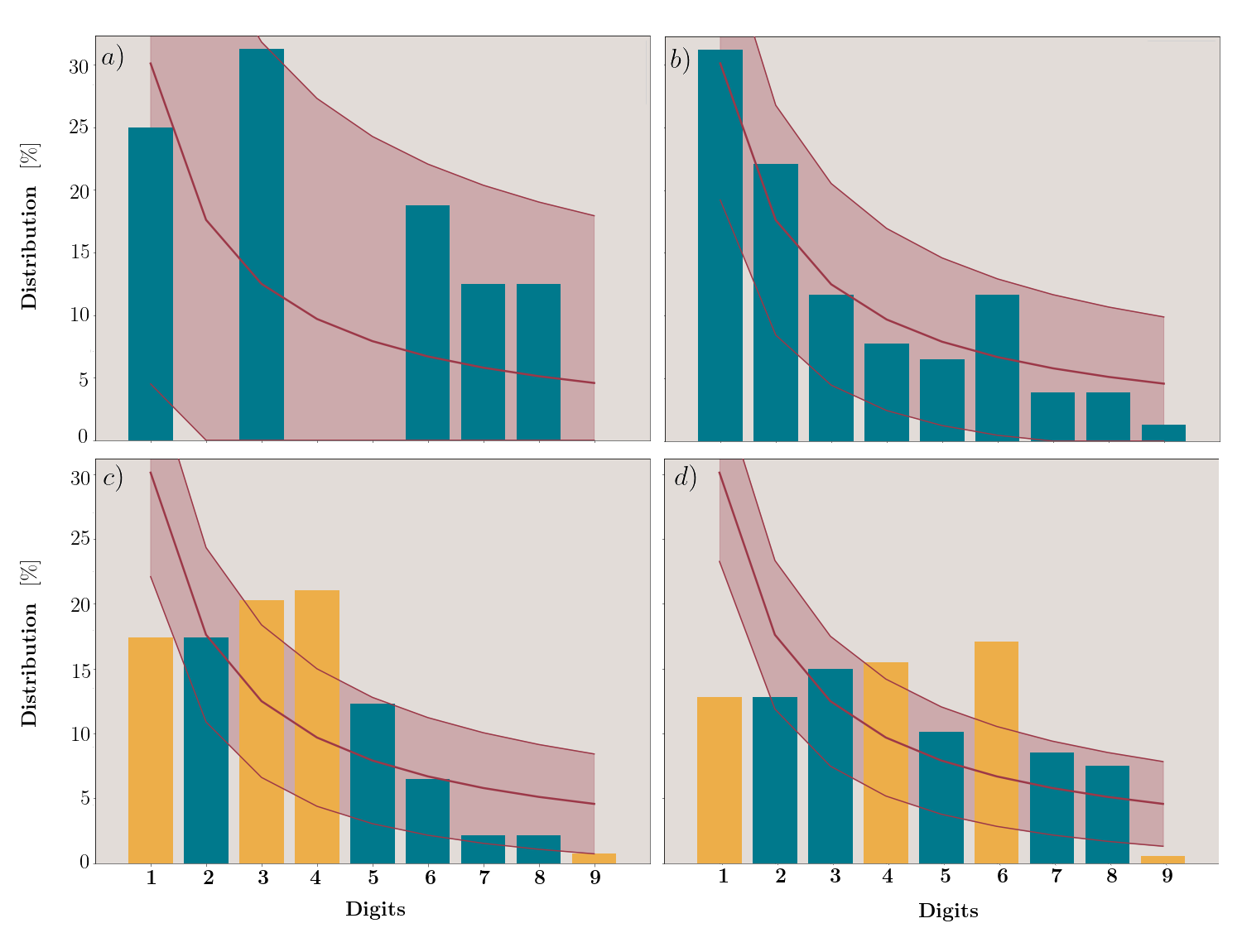}
      \label{fig:mean and std of net14-wavelet}
    \caption
    {The image \textbf{a} shows the distribution of first digits of Guarujá for the first period. Followed by \textbf{b}, the second, \textbf{c} the third and \textbf{d} the fourth.} 
    \label{guaruja}
  \end{figure}

\begin{figure}
    \centering
      \includegraphics[scale=1.3]{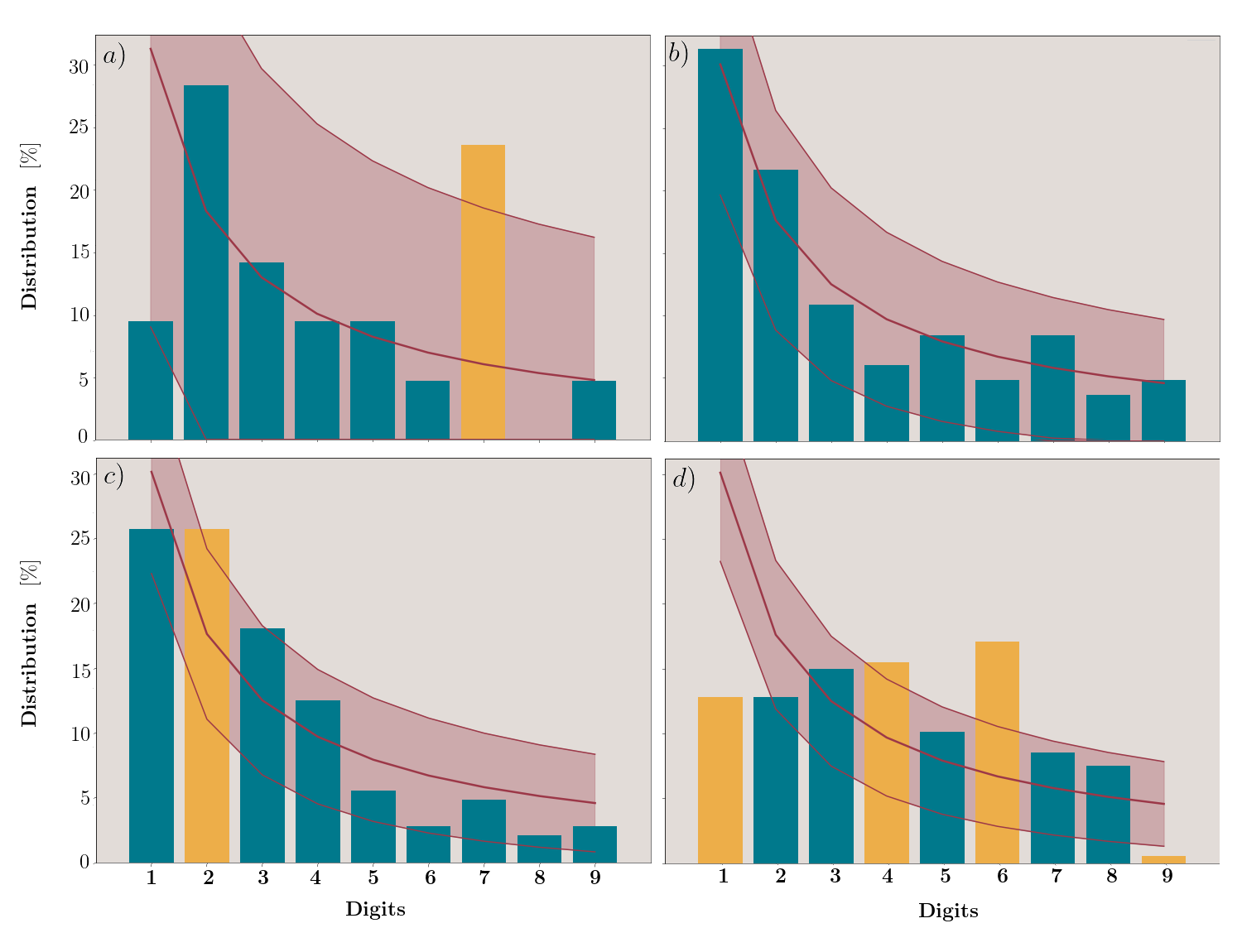}
      \label{fig:mean and std of net14-wavelet}
    \caption
    {The image \textbf{a} shows the distribution of first digits of Taboão da Serra for the first period. Followed by \textbf{b}, the second, \textbf{c} the third and \textbf{d} the fourth.} 
    \label{taboao}
  \end{figure}

\begin{figure}
    \centering
      \includegraphics[scale=1.3]{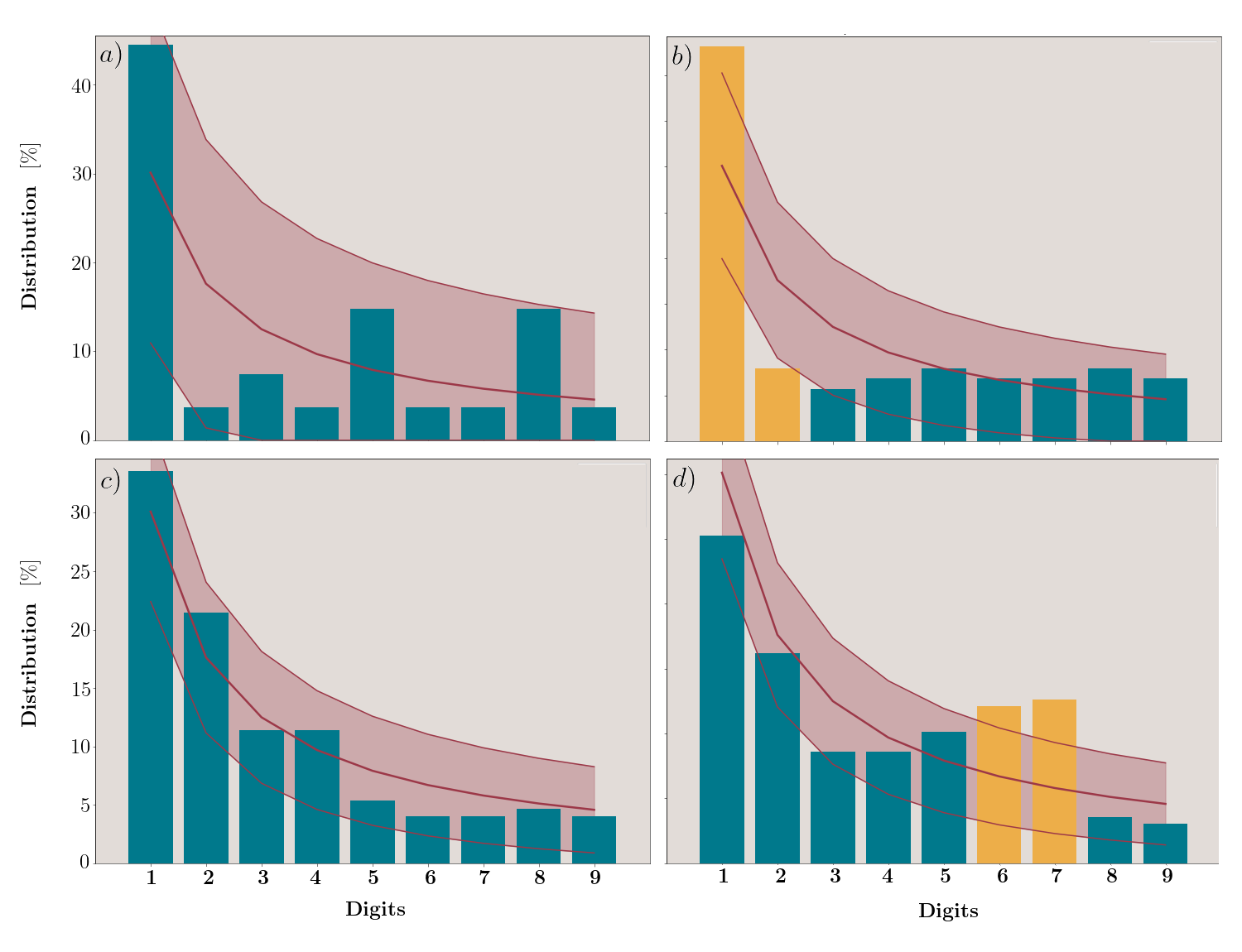}
      \label{fig:mean and std of net14-wavelet}
    \caption
    {The image \textbf{a} shows the distribution of first digits of Mogi das Cruzes for the first period. Followed by \textbf{b}, the second, \textbf{c} the third and \textbf{d} the fourth.} 
    \label{mogi}
  \end{figure}

\begin{figure}
    \centering
      \includegraphics[scale=1.3]{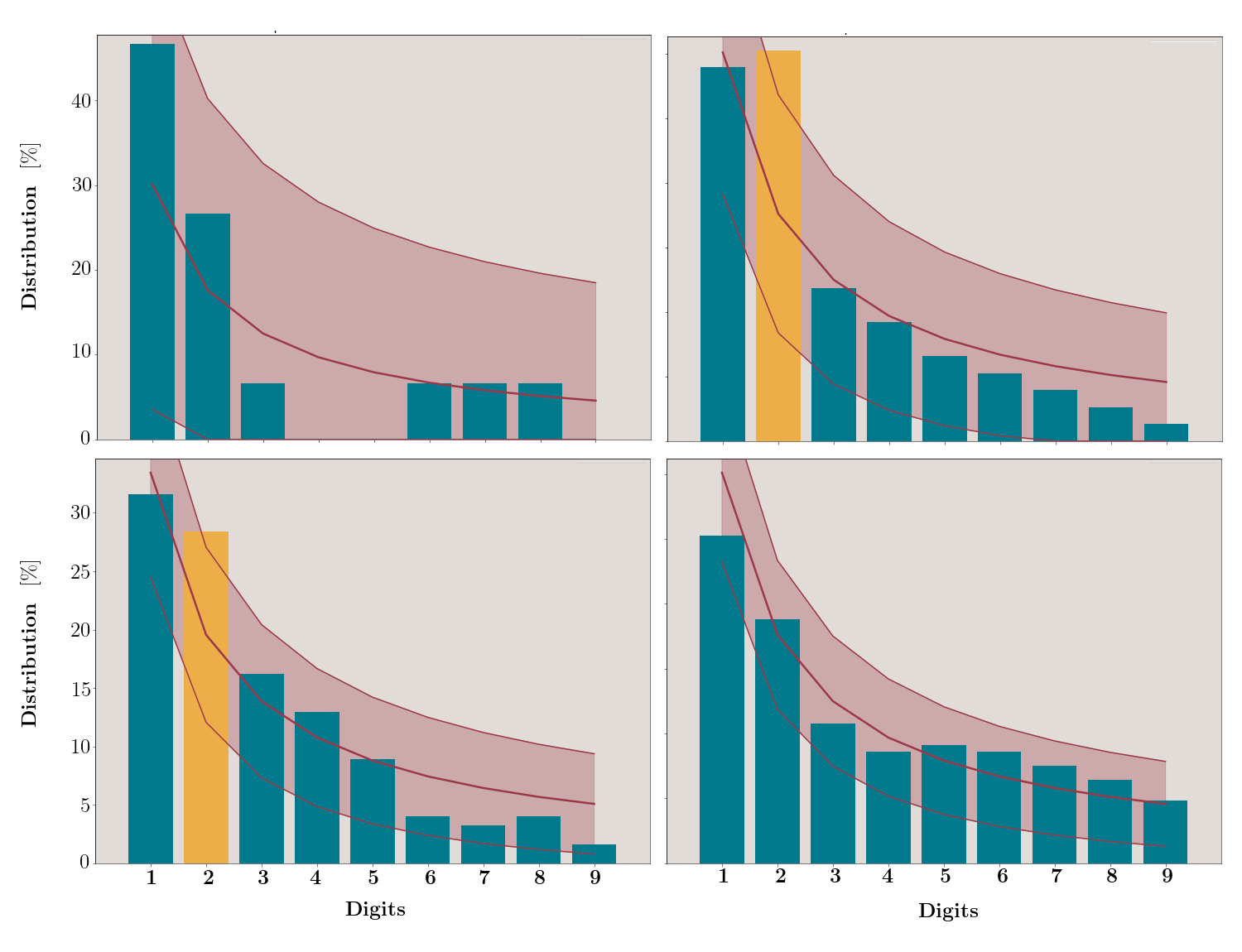}
      \label{fig:mean and std of net14-wavelet}
    \caption
    {The image \textbf{a} shows the distribution of first digits of Bauru for the first period. Followed by \textbf{b}, the second, \textbf{c} the third and \textbf{d} the fourth.} 
    \label{bauru}
  \end{figure}

In order to validate this concept, we performed the tests in four different situations: from February to April (first period), from February to June (second period), from February to August (third period) and from February to November (fourth period), as shown in figures \ref{guaruja}, \ref{taboao}, \ref{mogi}, \ref{bauru}. If the curve starts following Benford's law and subsequently starts to no longer follow it, then we can hypothesize that the curve has been flattened. However, if we obtain the opposite result, we are probably facing cases of underreporting that were being adjourned throughout the pandemic.

   These inconsistent results (between the BL test and the simulation table) are essential to note because they can discourage researchers from investigating any other self-reported data from the beginning of the pandemic further in detail. Besides, as reported by Silva \cite{SILVA}, an excess of lower digits is usually associated with underestimated cases of COVID-19. Under Benford's Law, we expect that mantissa is uniformly distributed with an average of $0.5$. Suppose figures are manipulated downward by exclusively altering the mantissa. In that case, the mean observed value will be less than the expected average.
  
  We observed that in the case of Guarujá, the numbers have only followed Benford's law in the beginning. This is a reasonably expected behavior since Guarujá is a coastal region with a lot of tourist activity. After several holidays and the lack of effective attitudes by the city, there was an extreme increase in cases. This increase generated an exponential growth, which can be seen in the second period. The last two periods shows numbers that are very different from the Benford distribution, which may mean that the measures to block the beaches and tourist centers were effective.

Moreover, Guarujá is a county with a high rate of testing, with this being proportionally higher than the rate of São Paulo state. This probably indicates that there is no underreporting in the third period and fourth periods. Besides, the mantissa means is $0.62$, which is higher than the expected average. This probably means that Guarujá did not manipulate the data, but maintained efficient containment measures. 
  
  Taboão da Serra, on the other hand, is a city with little tourist potential. In addition, for many years, the city was divided between the profile of a dormitory city and an industrial location. Until April, the social isolation rate varied from 48\% (during the week) to 58\% (on weekends) in the city. In the third period, this index dropped from 40\% (during the week) to 47\% (on weekends). Despite the quarantine, many businesses continued to function and, because it is still a dormitory city, the rate of social isolation remained low. On the other hand, for the last period, the mantissa means is $0.544$, which indicates that at least now, data is not being manipulated.
  
  Mogi das Cruzes showed a good index of social isolation in the second period, presenting 63\% (on weekends) and 51\% (during the week). However, this index has been decreasing. In addition, Mogi das Cruzes has almost 10\% of the population in rural conditions, which facilitates underreporting. Until July, deaths from severe acute respiratory syndrome without a definite cause increased more than 22-fold. 
  
  Bauru is a populous city in the midwest of the state of São Paulo. The testing rate is reasonably high in the region, so underreporting is not expected. On the other hand, few awareness actions were taken in the city to flatten the contagion curve, which can be observed with the recent increase in cases, in the so-called second wave.
 
Our findings are robust and increase our confidence that the epidemiological surveillance system fails to provide trustful data on the severe acute respiratory syndrome coronavirus 2 (SARS-CoV-2) epidemics in Brazil. There are two different hypotheses for the regions to follow or not Benford curve, which should be evaluated by sociologists and health experts. However, these results are significant for the creation of forecasting models, as we will present in the next section.
\subsection{Cases and Fatalities Forecasting}

The main objective of this section is to forecast the fatalities and daily new cases of COVID-19. One of the most famous approaches to forecasting is Seasonal Autoregressive Integrated Moving Average, SARIMA or Seasonal ARIMA (SARIMA). However, we have identified that some of the cities were not according to the Benford's distribution. In addition, every infectious disease outbreak exhibits specific patterns, and such patterns needed to be identified based on transmission dynamics of such outbreaks. Intervening measures to eradicate such infectious diseases rely on the methods used to evaluate the outbreak when it occurs. Likewise, forecasting models are related to the identification of patterns. If the standards are modified by containment measures or by low data quality, the model will not be highly accurate. Knowing if data is accurate is extremely important so that forecasts can be made more efficiently. 

Here, we consider the difficulty in identifying the 6 parameters necessary for the correct identification of the seasonal ARIMA for each of the 645 counties. We conducted tests with different cities of all clusters and observed that cities with similar sociodemographic characteristics also presented a similar behavior in relation to the pandemic, in proportionate terms. In this sense, table \ref{tab6} presents the base parameters that should be used for each of the clusters when forecasting daily new cases. 

\begin{table}[h!]
\begin{center}
\begin{tabular}{c|cc}
                & (p,d,q) & (P,D,Q) \\ \hline
Cluster 0       & (1,1,1)          &(1,1,1)            \\
Cluster 1       & (1,1,1)          &(0,1,1)             \\
Cluster 2       & (0,1,1)          &(0,1,1) \\
Cluster 3       & (1,1,1)          &(1,1,1)  \\
Cluster 4       & (0,1,1)          &(0,1,1)  \\
Cluster 5       & (0,1,1)          &(0,1,1)  \\
Cluster 6       & (1,1,1)          &(0,0,0)  \\
Cluster 7       & (1,1,1)          &(1,1,1)  \\
Cluster 8       & (0,1,1)          &(0,1,1)  \\
\end{tabular}
\end{center} 
\caption{Non-seasonal (p,d,q) and seasonal parameters (P,D,Q) of SARIMA to be used depending on cluster. We have defined the Akaike information criterion (AIC) as the estimator of prediction error and thereby relative quality metric of these statistical models. }\label{tab6}
\end{table}

Thus, in order to be able to predict these cases for each of the 645 counties, it would not be necessary to separately analyze the autocorrelations, partial autocorrelations, stationary conditions, etc. for each of them. It is only necessary to identify which cluster it belongs and follow the parameters defined in table \ref{tab6}. Adjustments can be applied to reduce residuals and AIC.

\begin{figure}
    \centering
      \includegraphics[scale=2]{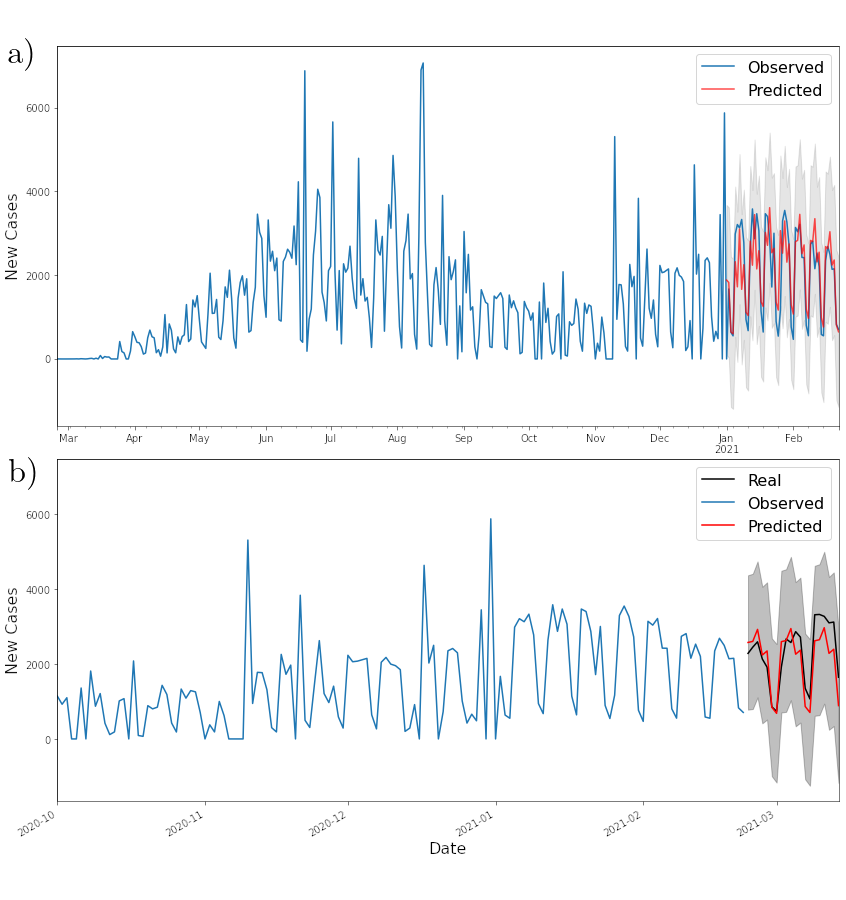}
      \label{fig:mean and std of net14-wavelet}
    \caption
    {The image \textbf{a} shows the forecast in São Paulo city considering the validation set,  which ranges from March of 2020 to end of February of 2021. The image \textbf{b} shows the forecast, in red, and the real values, in black, obtained using SARIMA model.} 
    \label{pred}
  \end{figure}

We conducted the forecast for 20 days using the capital, city of São Paulo, as an example. As it was segmented into cluster 2, we have chosen the parameters as (1,1,1)(0,1,1)[7]. The result of this forecast is shown in the figure \ref{pred}. The figure \ref{pred}.a presents the validation set, which ranges from March of 2020 to the end of February of 2021. Figure \ref{pred}.b shows the test dataset. 

\section{Conclusions}

This paper has investigated three conditions of São Paulo state counties regarding the pandemic of COVID-19. The first one concerns a clustering algorithm after the acquisition of 85 sociodemographic variables from the 645 municipalities in the state. Interestingly, these 645 counties generate 9 clusters with similar behaviors. Of these, four clusters have a high number of cases and fatalities, primarily due to the high demographic density, degree of urbanization and economic activity in retail and industries. 

As done in other countries, this type of analysis can facilitate restrictive measures to be taken in a coordinated manner taking into account the different regional characteristics. One downside factor regarding our methodology is that the addition of new features can modify the location of a given county in a cluster. Considering that other variables can be added to create a complete model, it is essential that hypotheses are raised \textit{a priori}, avoiding the addition of variables that cause collinearity.

Secondly, we have managed to apply Benford's law to estimate COVID-19 underreported data. To do so, we analyzed 4 counties with different characteristics during 4 different periods: from February to April (first period), from February to June (second period), from February to August (third period) and from February to November (fourth period). The conclusions of this section should be treated with caution. If the data does not follow Benford's law in any of the periods, it is quite possible that there is underreporting or manipulation of the data. Especially if the mantissa is less than the average expected in the Benford distribution.

However, an increasing change over the four periods could demonstrate a flattening of the contagion curve, which is expected by health authorities. Regardless of the situation, these inconsistent results are important to note because they can discourage researchers from investigating any other self-reported data by the specific county further in detail, such as by checking whether the hospitals are managing to cope with the number of infected patients admitted in critical care. As a result, BL testing alone would not be sufficient to detect potential manipulations of the growth of the death rate. For this reason, it is crucial to interpret epidemiological models that estimate parameters for the growth rate and deceleration of growth.

Finally, we worked with forecasting. We have used seasonal ARIMA models to predict daily cases in the counties. It was noticed that the ARIMA model parameters can be defined by the clusters. In this way, cities within the same clusters can have their daily cases data effectively predicted by the SARIMA model with the same parameters, optimizing the process. Thus, instead of defining $6 \times 645 = 3870$ parameters to forecast new cases from all the cities, we need only $6 \times 9 = 54$ to create models for all the counties.

In this sense, the framework proposed using these four tools is of paramount importance so that health and government authorities, not only from the state of São Paulo but from any state, city or country - as the methodology can be replicated in several granularities - are able to define strategies more effective data isolation, prediction and reliability. In addition, we demonstrate how different techniques of machine learning, deep learning, data mining and statistics can be used together to produce important results when dealing with pandemic data.

\section*{Funding}
This study was funded in part by the Brazilian funding agencies FAPESP -- S\~{a}o Paulo State Research Support Foundation (Grant No. 2019/15754-7).

\section*{Conflict of Interest}
The authors declare that they have no conflict of interest.


\end{document}